\documentclass[12pt]{mytmp}

\usepackage{fullpage}

\usepackage[utf8]{inputenc} 
\usepackage[T1]{fontenc}    
\usepackage{hyperref}       
\usepackage{url}            
\usepackage{booktabs}       
\usepackage{threeparttable} 
\usepackage{amsfonts}       
\usepackage{nicefrac}       
\usepackage{microtype}      
\usepackage{color,xcolor}
\usepackage{graphicx}
\usepackage{multirow}
\usepackage{enumitem}
\usepackage{listings}

\def\AA{\mathcal{A}}
\def\MM{\mathcal{M}}
\def\OO{\mathcal{O}}
\def\PP{\mathcal{P}}
\def\SS{\mathcal{S}}

\definecolor{dkgreen}{rgb}{0,0.6,0}
\definecolor{gray}{rgb}{0.5,0.5,0.5}
\definecolor{mauve}{rgb}{0.58,0,0.82}

\lstdefinestyle{mypython}{
  language=Python,
  aboveskip=1mm,
  belowskip=1mm,
  showstringspaces=false,
  columns=flexible,
  basicstyle={\small\ttfamily},
  numbers=none,
  numberstyle=\tiny\color{gray},
  keywordstyle=\color{blue},
  commentstyle=\color{dkgreen},
  stringstyle=\color{mauve},
  breaklines=true,
  breakatwhitespace=true,
  tabsize=3
}

\lstdefinestyle{mysh}{
  language=bash,
  aboveskip=1mm,
  belowskip=1mm,
  basicstyle={\small\ttfamily},
}

\journal{Artificial Intelligence}

\begin{document}

\begin{frontmatter}

\title{TLeague: A Framework for Competitive Self-Play based Distributed Multi-Agent Reinforcement Learning}

\author[ailab]{Peng Sun\corref{cor1}}
\ead{pengsun000@gmail.com}
\author[ailab]{Jiechao Xiong\corref{cor1}}
\ead{jchxiong@gmail.com}
\author[ailab]{Lei Han\corref{cor1}}
\ead{leihan.cs@gmail.com}
\author[ailab]{\\Xinghai Sun}
\author[thu]{Shuxing Li\fnref{intern}}
\author[thu]{Jiawei Xu\fnref{intern}}
\author[ailab]{Meng Fang}
\author[ailab]{Zhengyou Zhang}

\cortext[cor1]{Equal contribution, correspondence to the first three authors.}
\fntext[intern]{Work was done during the internship with Tencent Robotics X.}

\address[ailab]{Tencent Robotics X, Shenzhen, China}
\address[thu]{Tsinghua University, Shenzhen, China}

\begin{abstract}
Competitive Self-Play (CSP) based Multi-Agent Reinforcement Learning (MARL) has shown phenomenal breakthroughs recently. 
Strong AIs are achieved for several benchmarks,
including Dota 2, Glory of Kings, Quake III, StarCraft II, to name a few.
Despite the success, 
the MARL training is extremely data thirsty, 
requiring typically billions of (if not trillions of) frames be seen from the environment during training in order for learning a high performance agent.
This poses non-trivial difficulties for researchers or engineers and prevents the application of MARL to a broader range of real-world problems.
To address this issue,
in this manuscript we describe a framework, 
referred to as TLeague, 
that aims at large-scale training and implements several main-stream CSP-MARL algorithms.
The training can be deployed in either a single machine or a cluster of hybrid machines (CPUs and GPUs), 
where the standard Kubernetes is supported in a cloud native manner.
TLeague achieves a high throughput and a reasonable scale-up when performing distributed training.
Thanks to the modular design, 
it is also easy to extend for solving other multi-agent problems or implementing and verifying MARL algorithms.
We present experiments over StarCraft II, ViZDoom and Pommerman to show the efficiency and effectiveness of TLeague. 
The code is open-sourced and available at\\
\url{https://github.com/tencent-ailab/tleague_projpage}
\end{abstract}

\end{frontmatter}

\section{Introduction}
\label{sec:intro}
In the last decade, 
Deep Reinforcement Learning (DRL) was shown to be a powerful tool for sequential decision-making problems. 
Human or super-human level performance is achieved for several well-known benchmarks, 
including Atari~\cite{mnih2015human}, Go~\cite{silver2016mastering,silver2017mastering}, 
etc.
In particular, 
the Competitive Self-Play (CSP)~\cite{bansal2017emergent} based Multi-Agent Reinforcement Learning (MARL) has obtained impressive success for some non-trivial problems,
e.g., 
Dota 2~\cite{berner2019dota}, 
Glory of Kings~\cite{ye2020mastering},
Quake III~\cite{jaderberg2019human}, 
StarCraft II~\cite{vinyals2019grandmaster}.
The CSP-MARL method is Game Theoretic justified, 
as it performs \emph{Nash Equilibrium} (NE) finding by Fictitious Self-Play where the opponent mixture is stochastically approximated by opponent sampling and the \emph{Best Response} is realized with (Single-Agent) Reinforcement Learning that serves as a proxy algorithm~\cite{lanctot2017unified,balduzzi2019open,li2018sampled,swenson2017single}.
It is thus appealing in that the method is general-purpose and can be potentially applied to any MARL problem (Note that in cooperative or cooperative-competitive hybrid games, 
we can also sample for team-mates as in~\cite{jaderberg2019human}).
However, 
such a method is extremely data thirsty.
For example, 
in the aforementioned Dota 2 or StarCraft II,
it usually requires billions of (if not trillions of) frames generated from the environment during training (the equivalent in-game time is hundreds or thousands of years), 
otherwise the agent could be still under-trained and unlikely to perform well.
It thus poses prominent difficulties for RL researchers or practitioners when applying the method to their own interested problems.

To address this issue,
we propose a framework that aims at large-scale CSP-MARL training.
We adopt a modular design. 
An Actor-Learner-InferenceServer architecture~\cite{espeholt2018impala,espeholt2019seed} is taken to tackle the data producing (i.e., generating trajectories by interacting the environment and the agents) 
and the data consuming (i.e., learning from the trajectories by gradient descent).
There are also dedicated modules that manage an opponent pool and the concrete neural net parameters.
The modules coordinate to do the CSP-MARL training in parallel,
with each module having a high work-load and little idle time.
This way, 
it is able to maximally leverage a cluster of hybrid machines (with both CPUs and GPUs) and to substantially accelerate the training.
In our testing, it achieves a high throughput and a reasonable scale-up when using hundreds of GPUs and tens of thousands of CPU cores.
The large-scale run supports standard Kubernetes,
and the development with TLeague can be in a cloud-native manner. 

We have shipped with TLeague several main-stream algorithms .
The opponent sampling can be Population Based Training (PBT)~\cite{jaderberg2019human},
Agent-Exploiter~\cite{vinyals2019grandmaster},
(Prioritized) Fictitious Self-Play~\cite{berner2019dota,vinyals2019grandmaster}.
Several typical \emph{Policy Gradient} methods, 
such as PPO~\cite{schulman2017proximal} and V-trace~\cite{espeholt2018impala}, 
are supported.
One can also build desired neural nets in various architectures,
ranging from a simple list structure to a complicated \emph{Directed Acyclic Graph} (DAG).
The code is designed to be flexible and friendly to extend TLeague to other multi-agent problems (e.g., adding new environments, or RL algorithms, or opponent sampling algorithms, etc).

The rest of this manuscript is organized as follows.
We briefly review the related work in Section~\ref{sec:related-work}.
Then we describe CSP-MARL and explain the design of our code implementation in Section~\ref{sec:arch}.
Finally, we discuss several experiments over StarCraft 2, ViZDoom~\cite{kempka2016vizdoom} and Pommerman~\cite{resnick2018pommerman} in Section~\ref{sec:exp} to show the efficiency and effectiveness of TLeague.

\section{Related Work}
\label{sec:related-work}
Since the breakthroughs of DRL in Atari~\cite{mnih2015human} and GO~\cite{silver2016mastering,silver2017mastering},
several distributed RL frameworks were discussed in the literature.

Gorila~\cite{nair2015massively} decouples the Actor, Learner and Replay Memory to allow a scalable distributed training.
However, it only targets for DQN and uses asynchronous gradient descent.
Ape-X~\cite{horgan2018distributed} extends the Gorila framework in that it supports a centralized Prioritized Replay and using synchronous gradient descent.
Ape-X also incorporates more Q value based RL algorithms.
R2D2~\cite{kapturowski2018recurrent} is a successor of Ape-X, 
where the same architecture is adopted for implementing a more state-of-the-art Q learning. 

In another line of work,
attempts are made to parallelize the policy gradient algorithm.
A3C~\cite{mnih2016asynchronous} collects the trajectories and updates neural net gradients both in an asynchronous way, 
and only CPUs are used.
GA3C~\cite{babaeizadeh2016reinforcement} explicitly uses a GPU for neural net learning, 
and collects the trajectories across each actor asynchronously.
The method is efficient, 
but it can cause data lagging and henceforth hurt the performance of an on-policy RL.
To alleviate the problem, 
batched A2C~\cite{clemente2017efficient} forces a simultaneous environment stepping and collect the batch synchronously.
This method works for both on-policy and off-policy.
However, 
GA3C (or batched A2C) uses the same GPU for both forward-passing (collecting trajectories) and backward-passing (learning),
which can be still a bottleneck for scalable training.

IMPALA~\cite{espeholt2018impala} also uses GPUs and explicitly takes a decoupled Actor-Learner architecture,
where a state-of-the-art off-policy algorithm called V-trace is implemented.
SEED~\cite{espeholt2019seed} improves over IMPALA by performing the forward-pass over a separate Inference Server, 
which further increases the training speed.

The reverb library~\cite{reverb} implements a dedicated distributed Replay Memory,
which can be a building block for other RL framework.
The acme library~\cite{hoffman2020acme} builds on top of reverb,
adopting the Actor-Learner-InferenceServer architecture and implementing a bunch of modern RL algorithms.

The work mentioned above only addresses single agent RL and not covers MARL. 
In~\cite{jaderberg2019human,berner2019dota,vinyals2019grandmaster},
the authors discuss a proprietary framework for CSP-MARL,
and the code implementation is not publicly available.

The ray~\cite{moritz2018ray} library defines several \emph{primitives} for parallel computing,
on top of which an RL library called rllib~\cite{liang2018rllib} is built.
However, the TLeague framework discussed in this manuscript takes a design that is much closer to RL algorithms and to the machines.
No extra abstraction of parallel computing is introduced,
allowing TLeague be easy to run in cloud native manner (i.e., using Kubernetes).

Our work is most similar to IMPALA and SEED in regards of how it decouples RL components,
i.e., we also adopt the Actor-Learner-InferenceServer architecture.
However, we make several notable extensions.
1) We support CSP-MARL, for which separate modules are designed to manage an opponent pool and maintain the neural net parameters.
2) We use Horovod~\cite{sergeev2018horovod} to do the synchronous gradient updating across multiple GPUs for the Learners.

There is other work~\cite{castro18dopamine,tian2017elf,song2020arena} that devotes to providing various RL algorithms or environments,
which is beyond our scope.
However, we do borrow the code from the library openai/baselines~\cite{baselines} and deepmind/trfl~\cite{trfl} when implementing PPO~\cite{schulman2017proximal} and V-trace~\cite{espeholt2018impala} in TLeague.

\section{Architecture}
\label{sec:arch}

\subsection{Mathematical Settings}
\label{sec:math}
We presume the readers have been familiar with Single-Agent RL (cf. to, e.g., \cite{sutton1998introduction}).
Now we adopt the MARL settings from \cite{littman1994markov} and provide a brief description.
A Multi-Agent \emph{game} is indicated by a tuple $<\SS,\PP,r^i,\gamma,\OO^i,\AA^i,\pi^i>$ where $i \in \{1,2,...,N\}$ is the index for the $N$ agents.
The \emph{state} $s_t \in \SS$ fully describes the game at time step $t$.
The $o_t^i \in \OO^i$ and $a_t^i \in \AA^i$ are the \emph{observation} and \emph{action} for agent $i$, respectively. 
The dynamics of an \emph{environment} is carved by a transition $\PP:\SS \times \AA^1 \times \AA^2 ... \times \AA^N \mapsto \SS$ which is represented by a \emph{stationary} probability $P(s_{t+1}|s_t,a_t^1,a_t^2,...,a_t^N)$ over the next state $s_{t+1}$ when all the agents perform the actions $a_t^1, a_t^2,...,a_t^N$ at current state $s_t$. 
Denote by $r_t^i: \SS \times \AA^1 \times \AA^2 ... \times \AA^N \mapsto \mathbb{R}$ the instantaneous \emph{reward} $r_t^i (s_t, a_t^1, a_t^2,...,a_t^N)$ received for agent $i$ at time $t$.
The $i$-th agent's \emph{policy} $\pi^i: \OO^i \mapsto \AA^i$ gives the conditional probability $\pi^i(a_t^i|o_t^i)$ over the action $a_t^i$ to take when observing $o_t^i$.
The policy $\pi_{\theta}^i$ is usually represented by a parametrized \emph{function approximator}, 
e.g., a neural network with parameters $\theta$.
When the function approximator is able to model sequential data (e.g., with an LSTM layer~\cite{hochreiter1997long}),
we can let the policy be conditioned on the entire observation history and rewrite it as $\pi^i(a_t^i|\{o_{ t'}^i \}_{t'=0}^t)$.
An MARL algorithm seeks to learn, for each agent, an optimal policy $\pi^i$ that maximizes the expected \emph{return} $\mathbb{E}_{\PP,\pi^1,...,\pi^N}[\sum_{t=0}^T \gamma^t r_t]$, 
where $T$ is the \emph{horizon} (i.e., the episode length in time for the game) and $\gamma \in (0, 1]$ is the \emph{discount factor} that prevents the inflation for a too large or infinite horizon $T$.
In particular, 
a \emph{model-free} algorithm is able to learn $\pi^i$ when $\PP$ and $r^i$ are unknown in their mathematical forms.

Note that the reward structure implies the game \emph{mode}: is it competitive, cooperative, or hybrid?
For example, suppose a two-agent zero-sum game where we have $r^1 + r^2 = 0$. 
The game is then competitive, as one agent benefits from the other agent's loss. 
We note that various real-world games can be modeled by such a competitive mode, ranging from Rock-Paper-Scissor to StarCraft 1vs1 full game.

How to perform a concrete MARL training?
It is tempting to independently apply an (Single-Agent) RL for each agent.
However, 
this is not mathematically justified, 
as in this way the agent $i$'s dynamics is \emph{non-stationary} when the other agents' policies are absorbed into agent $i$'s environment. 
Such a treatment contradicts to the presumption of most model-free RLs that are derived for stationary dynamics.
There exist cases that independent RL is reported to be effective for MARL~\cite{vlassis2007concise},
but in many other applications it leads to poor results~\cite{lanctot2017unified}.
In particular, 
it suffers from the policy-forgetting during training when the policy space is rich and contains circulation~\cite{lanctot2017unified,balduzzi2019open}.
An example is the game Rock-Paper-Scissor. 
A naive independent RL will circulate over pure-rock, pure-paper, pure-scissor, ... that the late policy (e.g., pure-scissor) forgets how to beat the early policy (e.g., pure-rock). 

From the perspective of Game Theory, 
the ``gradient field'' of independent RL rotates over (but never converges to) an optimal point (i.e., the NE in parameter space). 
A remedy to this is the Fictitious Self-Play (FSP) algorithm that dates back to the 1950s~\cite{robinson1951iterative}.
Below we briefly explain FSP in a competitive two-agent case, 
where one is the learning agent and the other is an opponent.
During FSP training, 
the learning agent plays against a mixture of the historical opponents, 
not just the current opponent as in independent RL. 
This way, 
it introduces a ``centripetal force'' pointing to NE in the gradient field (e.g., see Fig. 4 of~\cite{srinivasan2018actor}), 
avoiding policy forgetting and converging to NE.

In the case of a multi-step game (e.g., StarCraft II), 
it is non-trivial to implement FSP in a straightforward way, 
as one should maintain a mixture of historical policies conditioned at \emph{every} state $s \in \SS$ (Note in a one-step game like Rock-Paper-Scissor, there is only one possible state and is usually omitted).
There are studies \cite{heinrich2015fictitious,heinrich2016deep} where a  neural network is adopted to model the conditional mixture term
\footnote{In a more general setting, 
it is possible to perform a no-regret learning for NE finding~\cite{waugh2014unified,burch2018time,johanson2007robust,zinkevich2008regret,brown2019deep,li2018double}. 
The corresponding discussion is beyond the scope of this manuscript.}.
However, a more straightforward and convenient implementation is the opponent sampling based Monte Carlo method. 
Denote by $\theta$ the parameter of an agent's policy. 
Construct a pool $\MM = \{\theta_1, \theta_2,...\}$.
On each episode beginning during training, 
an opponent, 
denoted by its parameter $\phi$, 
is selected by sampling from the pool $\phi \sim Q(\MM) $. 
Various sampling distributions $Q$ have been reported in the literature, 
including uniform~\cite{bansal2017emergent}, 
a probabilistic mixture of the current and the historic opponent~\cite{berner2019dota},
probabilistic Elo score matching~\cite{jaderberg2019human},
a function of win-rate~\cite{vinyals2019grandmaster},
etc.
The opponent sampling can be viewed as a stochastic approximation of the opponent policy mixture.

Once an opponent $\phi$ is selected, 
the parameter $\phi$ gets fixed and the learning agent tries to maximize the return by updating its own parameter $\theta$. 
In the Game Theory community, 
this procedure is referred to as Best Response, 
which is actually an RL in view of Machine Learning~\cite{lanctot2017unified}.
Note the fixed $\phi$ leads to a stationary opponent policy $\pi_{\phi}$, 
which is then absorbed into the environment and the dynamics remains stationary for the learning agent. 
To this extent, 
one can employ any favorite RL (e.g., PPO~\cite{schulman2017proximal}, V-trace~\cite{espeholt2018impala}, etc) as the ``proxy algorithm''.
Morden RL is able to learn from \emph{trajectory segment} defined as tuples of observation-reward-action in contiguous time steps:
\begin{equation}
	\tau = (o_t, r_t, a_t,o_{t+1},r_{t+1},a_{t+1},...,o_{t+L},r_{t+L},a_{t+L})
\end{equation}
where $L$ is the segment length and we've omitted the superscript for the learning agent.
This permits a mini-batch style SGD for RL which is more compatible to the Deep Learning paradigm.

Every once in a while, 
the pool is updated by $\MM \leftarrow \MM \cup \{ \theta \}$.
This way, 
the learning agent still plays against a mixture of historical opponents stochastically.
The initial size of the pool is one that $\MM = \{\theta_1\}$, 
where the ``seed'' policy parameter $\theta_1$ can be either randomly initialized or the one learned from \emph{Imitation Learning}.

Finally, we note that FSP is easy to extend to multiple opponents ($>= 2$).  
For example, 
do the sampling $\phi \sim Q$ for each of the opponents, respectively, 
on each episode beginning as in~\cite{jaderberg2019human}.

\begin{figure}[htbp]
\center
\includegraphics[width=0.8\linewidth]{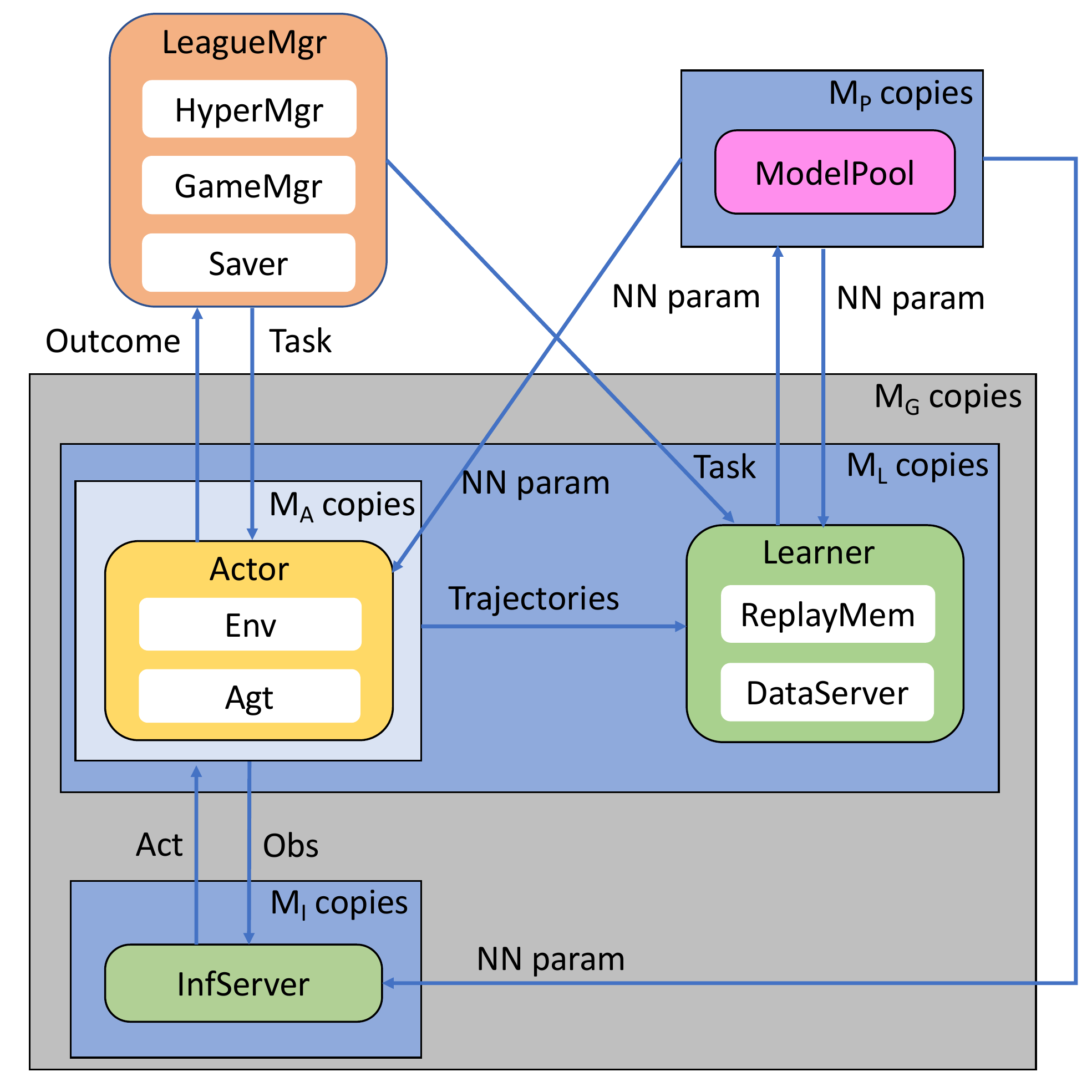}
\caption{
Diagram of the framework.
The rounded rectangle denotes a primary module or a secondary module (if any),
e.g., Actor is a primary module that embeds the secondary modules Env and Agt. 
Borrowing notations of \cite{jordan2004graphical}, 
we use a rectangle with a number on top-right to denote how many copies/replicas there are for a module.
In this convention,
we can read that there are $M_p$ ModelPools,
$M_A \times M_L \times M_G$ Actors,
etc.
The method-call (or message-passing) is represented by arrows.
The contact of the arrow indicates ``how the messages are packed''.
For example, 
the ``Trajectories'' arrow starts from the ``$M_A$ copies'' rectangle and ends at the ``Learner'' rounded rectangle, 
which indicates that the $M_A$ Actors altogether send trajectories to a single one Learner. 
See the text for detailed explanations.
}
\label{fig:diagram}
\end{figure}

\subsection{Design}
\label{sec:design}
To implement the CSP-MARL algorithm described in Section~\ref{sec:math} and allow it to be scalable,
we adopt a modular design for our distributed training framework.
Fig.~\ref{fig:diagram} gives an overview.
In the following,
we describe each of the modules and explain how they correspond to CSP-MARL.
\newline
\textbf{Actor.} 
The Actor module produces the trajectory for the learning agent.
It embeds two secondary modules, 
Env (environment) and Agt (agent).
We require Env be OpenAI gym~\cite{brockman2016openai} compatible for the Multi-Agent case,
that is, 
it should implement the two methods:
\begin{lstlisting}[style=mypython]
l_obs = env.reset()  # episode beginning
l_obs, l_rwd, done, info = env.step(l_act)  # in-episode stepping
\end{lstlisting}
where the \texttt{l\_obs} represents a list of the observations from all the $N$ agents $\{ o_0^i \}_{i=1}^N$ ($t=0$, episode beginning) or $\{ o_{t+1} \}_{i=1}^N$ ($t>0$, in-episode stepping). 
Likewise, 
\texttt{l\_rwd} means $\{ r_t^i \}_{i=1}^N$,
\texttt{l\_act} means $\{ a_t^i\}$,
\texttt{done} is a boolean variable indicating whether it is an episode ending,
\texttt{info} is a Python dictionary that holds the extra information you want to pass (For example, in StarCraft II environment we use \texttt{info['outcome']} to indicate if an agent wins/losses/ties the match of this episode).
The Agt carries a function approximator (usually a neural net) as the policy $\pi^i$ and takes the action $a_t^i \sim \pi^i$, 
where the neural net forward pass can be done either in a local machine or be delegated to a (remote) InfServer.
At each episode beginning, 
the Actor requests a \emph{task} from the LeagueMgr to know what the current (itself) learning policy $\pi_{\theta}$ and what the opponent policy $\pi_{\phi}$. 
At each episode ending, it also reports the game outcome to LeagueMgr.
During the Env-Agt interaction loop,
the trajectories of the learning agent are sent to Learner for the Neural Net training therein.
Periodically, the Actor pulls up-to-date policy parameters $\theta$ and $\phi$  from ModelPool.
\newline
\textbf{Learner.} 
A learning agent can own $M_L$ Learner modules.
Each Learner receives trajectories from $M_A$ Actors associated to it.
Each Learner can bind to a GPU,
and the $M_L$ Learners synchronize parameter gradients using the library Horovod~\cite{sergeev2018horovod} which performs an efficient \emph{allreduce} algorithm and can benefit from a fast inter-GPU connection via NCCL2~\cite{jeaugey2017nccl}. 
We allow there be up to $M_G$ learning agents that train in parallel.
In a full run, we have $M_G \times M_L$ Learners (henceforth that many GPUs) and $M_G \times M_A \times M_L$ Actors.
Each Learner embeds exactly one DataServer and one ReplayMem (Replay Memory),
performing a series of trajectory data pre-processing,
e.g.,
receiving the trajectory segments sent from actors and storing the data in a Replay Memory,
calculating the algorithm specific terms (say, the $\lambda$-return),
GPU-prefetching for the mini-batch to be learned,
etc. 
On each \emph{learning period} beginning, 
the Learner receives a \emph{task} from the LeagueMgr to know what the current policy $\theta$ it is training,
and the learner task must be consistent with the actor task.
During training, 
the Learner also periodically updates the policy parameter $\theta$ stored in the ModelPool.
Note the $M_L$ Learners are strictly synchronized,
thereby only one Learner suffices to do the task requesting.
We let the 0-th Learner (i.e., the \emph{rank}-0 machine in MPI~\cite{1999mpi} semantics) do the job.
At the end of a learning period, the current policy $\theta$ is frozen in the ModelPool.
A learning period should be long enough to ensure the policy has been well trained.
\newline
\textbf{InfServer.} 
The InfServer (abbreviation of Inference Server) is optional. 
When enabled, 
an InfServer collects a batch of observations from different Actors and feeds them into the neural net to predict the actions which are then returned to each of the Actors, respectively.
InfServer is usually deployed on GPU machines so that the batch forward-pass can be highly efficient.
Overall, such a scheme can lead to a higher throughput than that a one-step forward-pass (batch size 1) be done locally on each Actor.
In more sophisticated RL,
we may want to penalize the KL divergence between current policy and a teacher policy,
where we can also do the teacher policy forward-pass on an InfServer. 
\newline
\textbf{ModelPool.} 
The ModelPool stores the concrete neural net parameters of the opponent pool $\MM$.
During the whole training lifecycle,
ModelPool must respond to any parameter requesting (read) or updating (write) instantaneously, 
from either a Learner or an Actor.
We then use a load-balance technique for high concurrency and high performance,
where up to $M_M$ ModelPool replicas can be run simultaneously and a random one is picked to do the concrete response.
We also keep the neural net parameters in-memory to allow a fast read/write operation.
\newline
\textbf{LeagueMgr.} 
The LeagueMgr (abbreviation of League Manager) sponsors the training and coordinates the other modules.
A key secondary module is the GameMgr (Game Manager),
which maintains a payoff matrix for all the models stored in the pool $\MM$ and implements (in the derived class) various opponent sampling algorithms aforementioned in Section~\ref{sec:math}.
The selected learning agent and/or opponent is wrapped as a task sent to Actor or Learner.
Another secondary module is the HyperMgr (Hyper-parameter Manager), 
which maintains the hyperparameters associated with each model $\{\theta_i\} \in \MM$.
Here the hyperparameters can account for various algorithmic settings,
e.g., 
the learning rate or discount factor for RL, 
the variance term of the Gaussian Elo matching probability for opponent sampling~\cite{jaderberg2019human},
the z-statistics as required by the AlphaStar policy neural net~\cite{vinyals2019grandmaster},
etc.
The HyperMgr can also perturb or vary the hyper-parameters, 
as required by some algorithm like PBT~\cite{jaderberg2019human}.

\subsection{System-Level Design}
\label{sec:sys-desing}
For small-scale training,
we can simply run all the aforementioned modules in a single machine.
However, for contemporary large-scale RL that requires high throughput training,
it is unrealistic to do the single machine.
For example, one may need thousands of Actors to produce the trajectories in parallel and hundreds of GPU Learners to consume and learn from the data,
which exceeds the capacity of a common machine. 
Keeping this in mind,
we have designed the framework so that the modules can be deployed across multiple machines to support a scalable distributed training.
We adopt a \emph{Microservices} paradigm~\cite{balalaie2016microservices},
that is, 
each module can be launched as an OS \emph{Process},
exposing its APIs to other modules and behaving like a \emph{Service}.
The modules/processes talk to each other via \emph{RPC (Remote Procedure Call)},
which maps the code-level class interaction to system-level inter-process communication.
We define our private inter-process message (i.e., the API protocol) in native Python3 language and rely on the library ZeroMQ~\cite{zmq} for RPC.
\footnote{Other scheme is possible, e.g, using protobuf~\cite{protobuf} and gRPC~\cite{grpc}}

\subsection{Large-scale Run and Kubernetes}
When performing large-scale training, 
it is challenging to do it over \emph{bare metal} machines due to several reasons:
\begin{itemize}[noitemsep,nolistsep]
\item It is difficult and error-prone to initiate and maintain the basic setups for each machine, e.g., a list of the IPs/hostnames, password-less SSH, package dependencies, etc.
\item It is tedious to manually start/stop each module on the desired machine.
\item The error-tolerance mechanism is absent when running the modules. For example, we hope the Actor can auto restart when it crashes due to some low-level error that is out of our control (e.g., the occasional core-dump of the Env binary).
\item It is error-prone when updating and keeping consistent the code over all machines during the development period.
\end{itemize}
To address these issues,
we embrace the cloud-native philosophy~\cite{balalaie2016microservices}. 
We rely on Kubernetes (abbreviated as k8s)~\cite{luka2017kubernetes} to manage our large-scale distributed run
\footnote{We also provide examples of how to write Shell scripts for running in a single machine (or several machines) to cover the use-case of small-scale training.}
as follows.

Depending on the role it plays in the framework,
each TLeague module is made as a proper k8s \emph{resource}~\cite{luka2017kubernetes}.
Specifically, LeagueMgr, ModelPool, Learner and InfServer is made as k8s \emph{Service},
respectively, 
exposing the APIs via an endpoint idiom in the format of hostname:ip-port, 
(e.g., \texttt{tcp://signature-league-mgr:9003}).
Actor is made as k8s \emph{Deployment} or \emph{ReplicaSet},
which allows us to scale-up/-down the number of Actors to adjust the trajectory producing speed during the whole training lifecycle.
It also automatically restarts the Actor in case it encounters an unrecoverable error thanks to the k8s imperative semantic.
The concrete TLeague module carrier is a k8s \emph{pod}, 
which can be placed and co-located in desired type of machine using k8s \emph{nodeSelector}.
We prepare everything of a distributed training in a yaml file,
including both the RL algorithm settings (e.g., learning rate, discount factor, etc.) and the k8s settings.
Then we submit it to a k8s cluster using the kubectl command-line.
We also employ jinja2 (a template library~\cite{jinja2}) to generate the yaml in a configurable and concise way.
Suppose, for example, a training specification has been written in a file named \texttt{foobar.yml.jinja2},
then we can start or stop the training by simply running something like the bash commands: 
\begin{lstlisting}[style=mysh]
# start
python render_template.py foobar.yml.jinja2 | kubectl apply -f -
# stop 
python render_template.py foobar.yml.jinja2 | kubectl delete -f -
\end{lstlisting}
We note that an alternative solution is the kubeflow~\cite{kubeflow},
where we can write a dedicated ``operator'' for each TLeague module.
However, we found the yml-jinja2-kubectl way suffices and is convenient enough in all our experiments,
and will not discuss the kubeflow solution in this manuscript.

In this way, our specification of a distributed training becomes more compute-resource centric.
For example, a yml.jinja2 file can be as descriptive as plain English like
``Alright, I want 56 Learners and 8 InfServers, 
each Learner corresponds to 16 actors. 
Each Learner requires 1 GPU, 
each InfServer requires 1 GPU and each Actor requires 4 CPU cores. 
Every 7 Learners and 1 InfServer must be co-located in the same GPU machine. 
Each ModelPool must be placed in the machine with a high-speed Network Adaptor and >200G memory.''

TLeague can be run over any standard k8s cluster.
For example, most experiments in Section~\ref{sec:exp} were run over Tencent Cloud~\cite{tencent-cloud},
where a k8s cluster is created and maintained via TKE (Tencent Kubernetes Engine~\cite{tke}) that provides an off-the-shelf k8s solution.
The \emph{node} machines (most of them are Tencent Cloud CVM~\cite{cvm}) can be added/removed by several clicks through a web console (or by programming with Tencent Cloud APIs if you like).
Tencent Cloud supports \emph{Cluster Autoscaler} for those node machines.
In addition with a flexible on-demand pricing strategy,
it allows us to achieve a very high peak computing ability with limited (monetary) budget,
which is, we argue, the best solution to running a full-scale experiment that requires tremendous computing resources (e.g., hundreds of GPUs and tens of thousands of CPU cores) for only a few days or weeks.
Indeed, in most of the time during a development cycle, 
we should run small-scale experiments for idea verification or prototyping.

\begin{figure}[htbp]
\center
\includegraphics[width=0.98\linewidth]{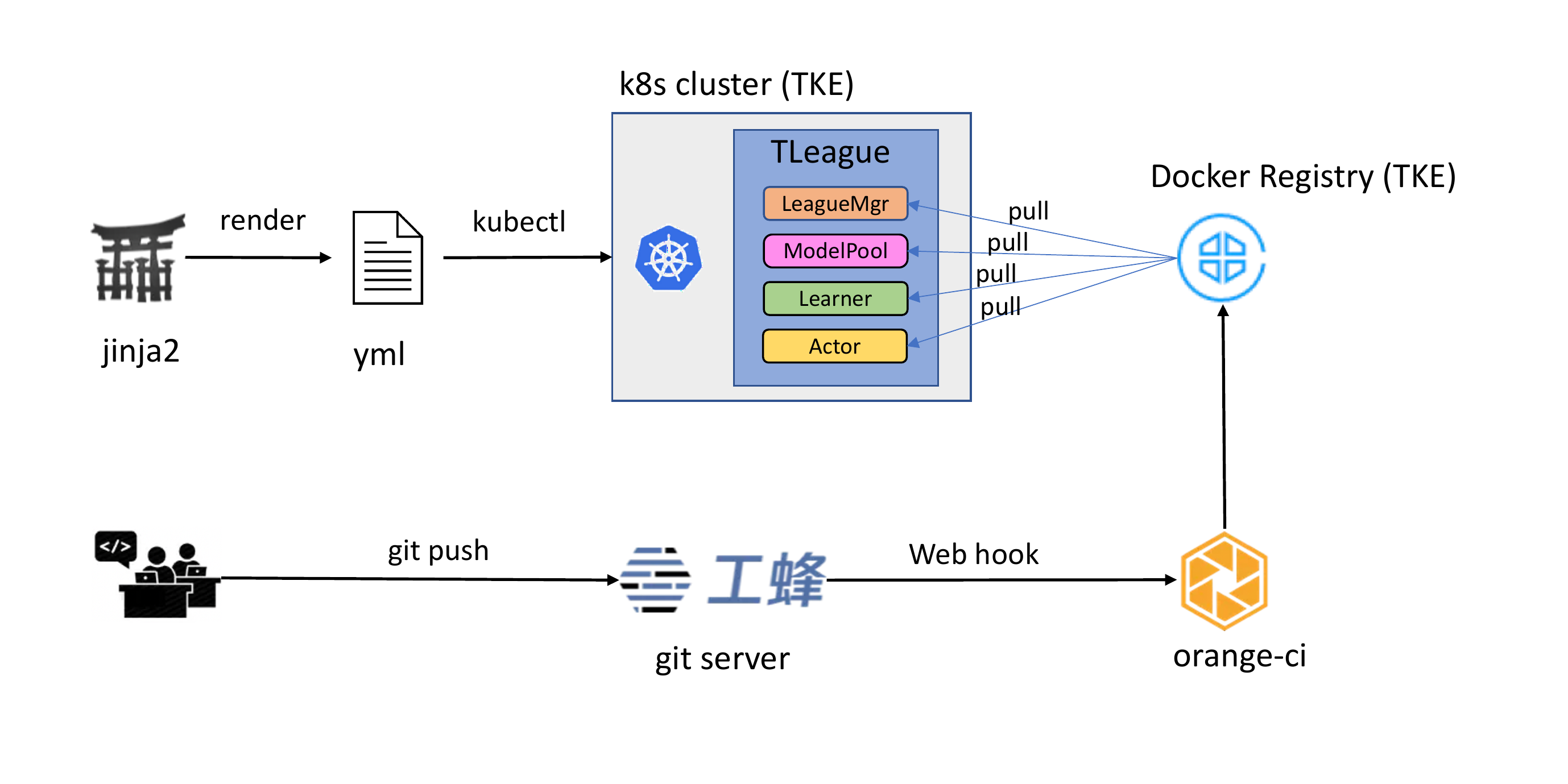}
\caption{
Workflow of our TLeague developing and training. 
See the text for detailed explanations.
Although some tools are currently Tencent internal, 
they have public available counterparts.
The CI tool Orange-CI can be replaced with Travis-CI.
Also, the private docker registry is usually provided by any other Cloud vendor 
(or just use the public dockerhub).
}
\label{fig:workflow}
\end{figure}

In our development, we also benefit from the Tencent Cloud ecosystem that provides abundant facilities.
We employ the CFS~\cite{cfs} (Tencent's extension to the standard NFS) and make it a k8s \emph{PVC (Persistent Volume Claim)} to do the directory sharing across pods.
We take advantage of the \emph{DevOps} concept to accelerate our developing iteration.
Specifically,
we use a CI (Continuous Integration) tool to build the \emph{Docker} \emph{Image} and push it to our private \emph{Docker Registry},
which can be triggered, for example, by a simple \emph{git} \emph{commit} pushing.
The full development workflow is summarized in Fig.~\ref{fig:workflow}.

\subsection{Code Structure}
\label{sec:code}
We've split the framework described in Section~\ref{sec:design} into several repositories in regards of the functionality,
satisfying a principle of high cohesion and low coupling.
Each repository is made as an independent Python package that can be pip installed.
In the following we provide more details for each repository.
\newline
\textbf{Arena.}
We've prepared the OpenAI gym compatible environments in a toolbox called \emph{Arena}~\cite{wang2019arena}. 
Moreover, the \emph{observation space} and \emph{action space} is implemented as \emph{Arena Interface},
which allows us to write the code only once for either training (binds to an environment like a \emph{gym env wrapper}) or testing (binds to an agent), 
see~\cite{wang2019arena} for detailed explanations of this mechanism.
When the logic of a specific environment is too heavy,
we recommend placing the corresponding code in a separate repository.
For example, 
we've made a repository called \textbf{TImitate} dedicated to SC2 full game, 
which involves feature engineering, replay file parsing, z-statistic extraction, etc.
\newline
\textbf{TPolicies.}
A Neural Network library tailored for Reinforcement Learning and Imitation Learning. 
It uses Tensorflow 1.x APIs and is in the style of the library \emph{tf-slim} and \emph{tf.contrib.layers}.
With TPolicies one can build policy net or value net in various architectures,
ranging from a simple one in list structure (e.g., a ConvNet plus LSTM for Atari~\cite{mnih2015human,mnih2016asynchronous} or ViZDoom~\cite{kempka2016vizdoom}) to a complicated one of general Directed Acyclic Graph (e.g., the net for SC2 full game~\cite{vinyals2019grandmaster}, containing layers/blocks of ResNet, Transformer, Pointer Net, Gated Linear Unit, Auto-regressive Action Heads, etc.).
TPolicies also provides RL related Tensorflow \emph{ops},
e.g., 
for building policy gradient loss, 
for computing $\lambda$-return.
Such RL related code is borrowed and adopted from the library openai/baselines~\cite{baselines} and deepmind/trfl~\cite{trfl}.
\newline
\textbf{TLeague.}
Most modules and the main functionality described in Section~\ref{sec:design} are implemented in this repository,
which depends on (must \texttt{import} in Python) Arena and TPolicies.

\subsection{Extension}
\label{sec:extension}
Thanks to the modular design, 
it's convenient to extend the framework for your own interested applications.
In the following we provide guidelines for some typical use-cases.
\newline
\textbf{Adding New Env.}
Prepare the Env code in \texttt{arena.env}, and write a thin wrapper in \texttt{tleague.envs}.
To customize the observation space and action space, 
write a corresponding Arena Interface~\cite{wang2019arena}.
Refer to the following Python modules for how to add the environment named pong-2p~\cite{wang2019arena}:
\begin{lstlisting}[style=mypython]
arena.env.Pong2pEnv  # game logic 
tleague.envs.pong # a thin wrapper to make it visible to TLeague
\end{lstlisting}
\textbf{Adding New RL Algorithm.}
Derive from \texttt{tleague.actors.BaseActor} for how to produce the trajectories (data producing), 
and derive from \texttt{tleague.learners.BaseLearner} for how to learn from the trajectories (data consuming).
Also, 
derive from \texttt{tleague.utils.DataStructure} to specify the trajectory data layout (e.g., a time step should contain an observation, a reward, a discount factor, etc.), 
serving as a contract between Actor and Learner.
Write the policy gradient related or value related loss in \texttt{tpolicies.losses}.
For example,
refer to the following Python modules for how to implement the V-trace~\cite{espeholt2018impala} algorithm:
\begin{lstlisting}[style=mypython]
tleague.actors.VtraceActor # Trajectories generating for V-trace 
tleague.learners.VtraceLearner  # Trajectories learning for V-trace
tleague.utils.VtraceData  # data layout for V-trace
tpolicies.losses.vtrace_loss. # V-trace related loss implementation
tpolicies.net_zoo.mnet_v6d6.mnet_v6d6.mnet_v6d6_loss  # use V-trace loss when building the net
\end{lstlisting}
\textbf{Adding New Opponent Sampling Algorithm.}
Derive from the class \\
\texttt{tleague.game\_mgr.GamgeMgr} and implement the required methods such \texttt{get\_player()}, \texttt{add\_player()}.
Refer to the following Python module as a minimal example which implements a uniform sampling from the historical opponents.
\begin{lstlisting}[style=mypython]
tleague.game_mgr.SelfPlayGameMgr
\end{lstlisting} 

\section{Experiment}
\label{sec:exp}
\begin{figure}[htbp]
\center
\includegraphics[width=0.35\linewidth]{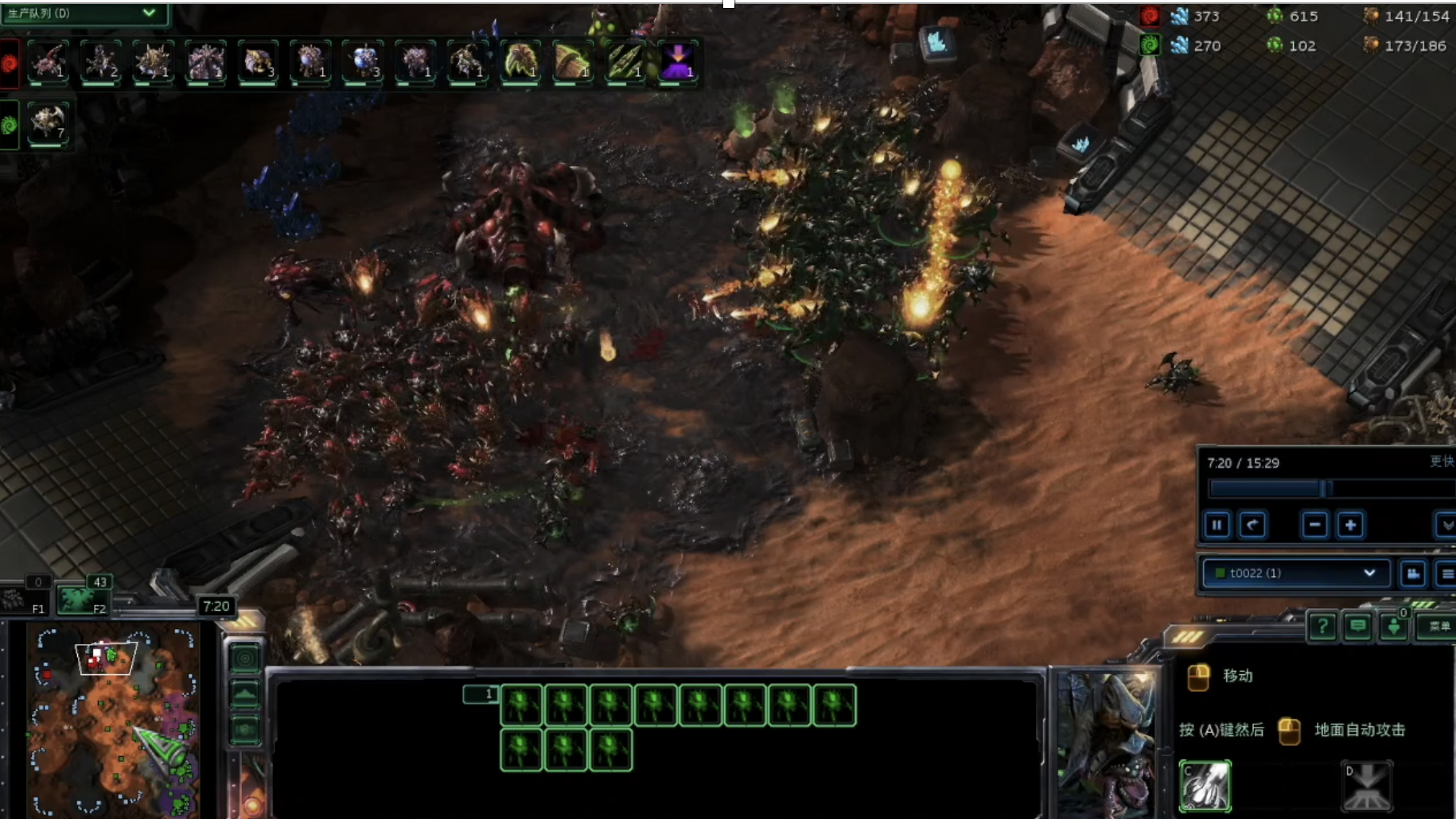}
\includegraphics[width=0.27\linewidth]{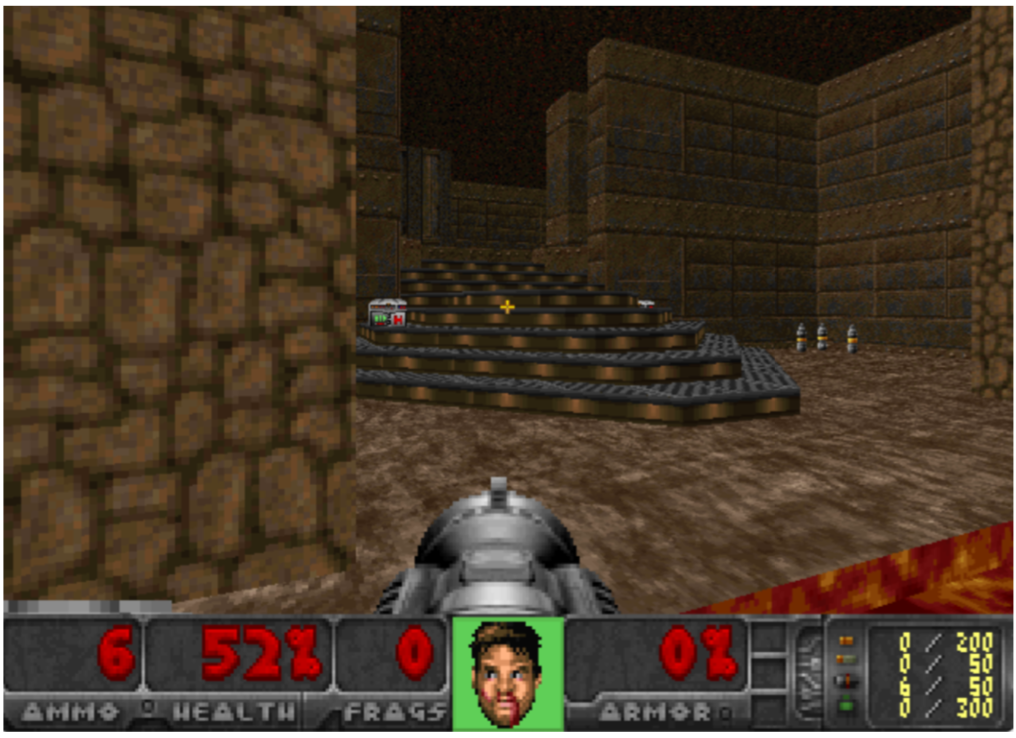}
\includegraphics[width=0.27\linewidth]{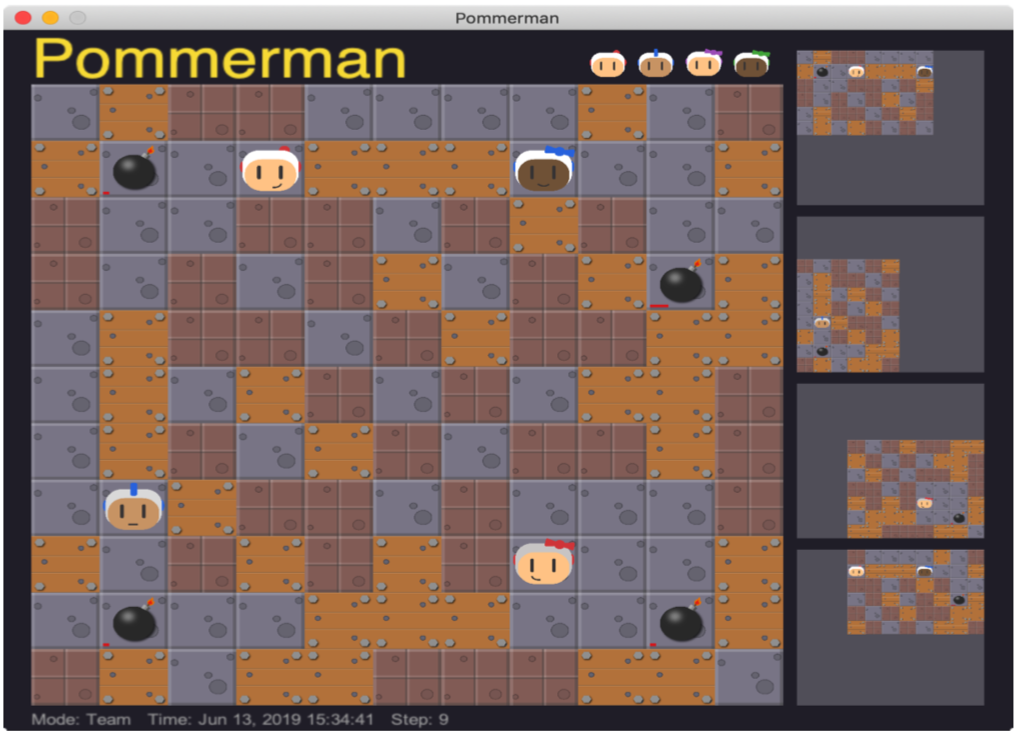}
\caption{
The games on which we did experiments: StarCraft II (left), ViZDoom (Middle) and Pommerman (Right).
}
\label{fig:games}
\end{figure}

We did experiments over three games: StarCraft II, ViZDoom and Pommerman (Fig.~\ref{fig:games}).

\subsection{SC2 full game}
\label{sec:exp-sc2}
We investigate StarCraft II zerg-vs-zerg full game and perform a CSP-MARL training with TLeague.
The technical details and the results are reported in a separate study~\cite{han2020tstarbotx}.

\subsection{ViZDoom}
\label{sec:exp-vizdoom}
ViZDoom~\cite{kempka2016vizdoom} is an AI research platform based on the FPS (First Person Shooter) game Doom.
We adopt the CIG 2016 competition track 1 protocol~\cite{cig2016},
where 8 AI players join in a maze and play against each other.
After a period of 10 minutes (in-game time),
the players are ranked by the FRAG, 
which is defined as kills minus suicides (due to own rocket splash).

In our experiment,
the observation is an RGB image, 
which is the first-person-view raw screen pixels as what a human player sees.
The action is discrete in the size of 6,
representing ``turn-left'', ``move-forward'', ``fire'', etc.
We employ a neural network consisting of 2 blocks of convolution layer followed by max pooling layer and an LSTM block.
We perform a two-stage training.
In the first stage,
the agent is trained to navigate in the maze,
for which we use reward shaping to encourage the agent to explore the map with the ``fire'' action disabled.
In the second stage,
the agent is trained by CSP-MARL for the full match, 
explained as follows.
On each episode beginning,
we sample from the pool $\MM$ for the rest 7 agents.
We simply adopt a uniform sampling over the most recent 50 models.
The proxy RL algorithm used in our experiment is PPO~\cite{schulman2017proximal}, 
where we've controlled the trajectory producing and consuming speed to ensure the on-policy.

The final trained agent (called \emph{MyPlayer} hereafter) is tested in several ways.
In Table~\ref{tab:myplayer-vs-bot} we give the results for playing against 7 builtin bots (called \emph{bots} for short hereafter).
\begin{table}[htbp]
  \centering
  \caption{A testing of 5 matches for ``1 MyPlayer, 7 bots''. FRAG is reported. In all the 5 matches, MyPlayer ranks 1.}
  \label{tab:myplayer-vs-bot}
  \begin{tabular}{l|c|c|c|c|c|c}
    \hline
             & 1  & 2  & 3  & 4  & 5  & Average \\
    \hline\hline
    MyPlayer & 26 & 24 & 31 & 27 & 30 & 27.6 \\
    \hline
  \end{tabular}
\end{table}

We also add a Single-Agent RL based AI named \emph{F1}~\cite{wu2017}, 
which was the champion for CIG 2016 track 1.
We test the following settings:
``1 MyPlayer + 1 F1 + 6 Builtin Bots'', 
``2 MyPlayer + 2 F1 + 4 Builtin Bots'',
``4 MyPlayer + 4 F1'',
and give the results in Table~\ref{tab:f}.
\begin{table}[htbp]
  \centering
  \caption{A testing of 5 matches for three settings. ``1 MyPlayer, 1 F1, 6 bots'' (top part), ``2 MyPlayer, 2 F1, 4 bots'' (middle part), ``4 MyPlayer, 4 F1'' (bottom part). The best FRAG is reported. For example, in the last row 29 is the best score of the 4 F1s in the 5th match.}
  \label{tab:f}
  \begin{tabular}{l|c|c|c|c|c|c}
    \hline
             & 1  & 2  & 3  & 4  & 5  & Average \\
    \hline\hline
    MyPlayer & 36 & 32 & 37 & 31 & 34 & 34.0 \\
    \hline
    F1       & 28 & 33 & 32 & 22 & 32 & 29.0 \\
    \hline\hline
    MyPlayer & 38 & 31 & 37 & 28 & 35 & 33.8 \\
    \hline
    F1       & 30 & 26 & 26 & 31 & 34 & 29.4 \\
    \hline\hline
    MyPlayer & 38 & 34 & 33 & 33 & 31 & 33.8 \\
    \hline
    F1       & 33 & 28 & 26 & 24 & 29 & 28.0 \\
    \hline
  \end{tabular}
\end{table}
The testing code and the ``F1'' policy net is adopted from~\cite{vizdoom-testing-code}.
All testing is done in a 12-core CPU desktop machine.
To ensure fairness,
we use the \emph{synchronous} mode for MyPlayer, F1 and the dedicated host so that the ViZDoom game core waits until it receives the actions from all players when performing in-game stepping.

Note that it is a self-play from scratch,
and the agent has never seen builtin bot or F1 during training. 
As can shown in the Tables, 
MyPlayer gets higher score (the FRAG) than both the builtin bot and F1.

\subsection{Pommerman}
\label{sec:exp-pommerman}
Pommerman~\cite{resnick2018pommerman} is a variant of the famous game Bomberman 
and is used as a benchmark for multi-agent learning. 
Typically, 
there are 4 agents that each can move and place bomb on an $11 \times 11$ board. 
At each step, 
an agent can take one of the 6 actions: \{Idle, Move Up, Move Down, Move Left, Move Right, Place a Bomb\}. 
A power-up item might appear when a wooden wall is destroyed by a bomb.  
Pommerman supports three modes: Free-for-All (FFA), Team and Team-Radio. 
In the FFA mode, 
the board is fully observable and each agent’s goal is to be the last survivor.
While in Team and Team-Radio mode, 
two agents cooperate as a team and fight against another 2-agent team.
Each agent can only see a $9 \times 9$ board in its neighborhood.
In Team mode agents are not allowed to communicate,
while in Team-Radio mode a limited-bandwidth radio between teammates is provided.
The team wins if it eliminates the 2 agents in the opponent team, 
and it gets a tie if the game is not finished within a maximal length of 800 steps.

In our Pommerman experiments, 
we adopt the Team mode, 
which is also used as the NeurIPS 2018 Competition environment. 
We use a decentralized policy to control the two agents. 
Each agent only uses its own observation, 
including the fogged board and self's attributes indicating the ammo number, 
blast strength, 
can-kick power-up and alive. 
The attributes are expanded as constant-value images that are concatenated with the board channels, 
yielding the feature maps as the observation. 
We employ a neural network consisting of 5 convolutional blocks,
followed by a gather op to collect the feature vector from the pixels where the agents reside, then passed through an LSTM block. 
To encourage the cooperation between the two teammate agents, 
we build a centralized value network that takes as inputs the two agents' LSTM embeddings.
The output value is then used as the critic for PPO~\cite{schulman2017proximal}. 
The two-agent team is viewed as a single agent (by performing the forward-pass twice for the only one neural net), 
and is trained from scratch with CSP-MARL.
The opponent sampling is a mixture of 35\% pure self-play and 65\% PFSP,
which is much like how the \emph{Main Agent} samples as described in~\cite{vinyals2019grandmaster}.

\begin{figure}[ht]
    \centering
    \includegraphics[width=0.45\textwidth]{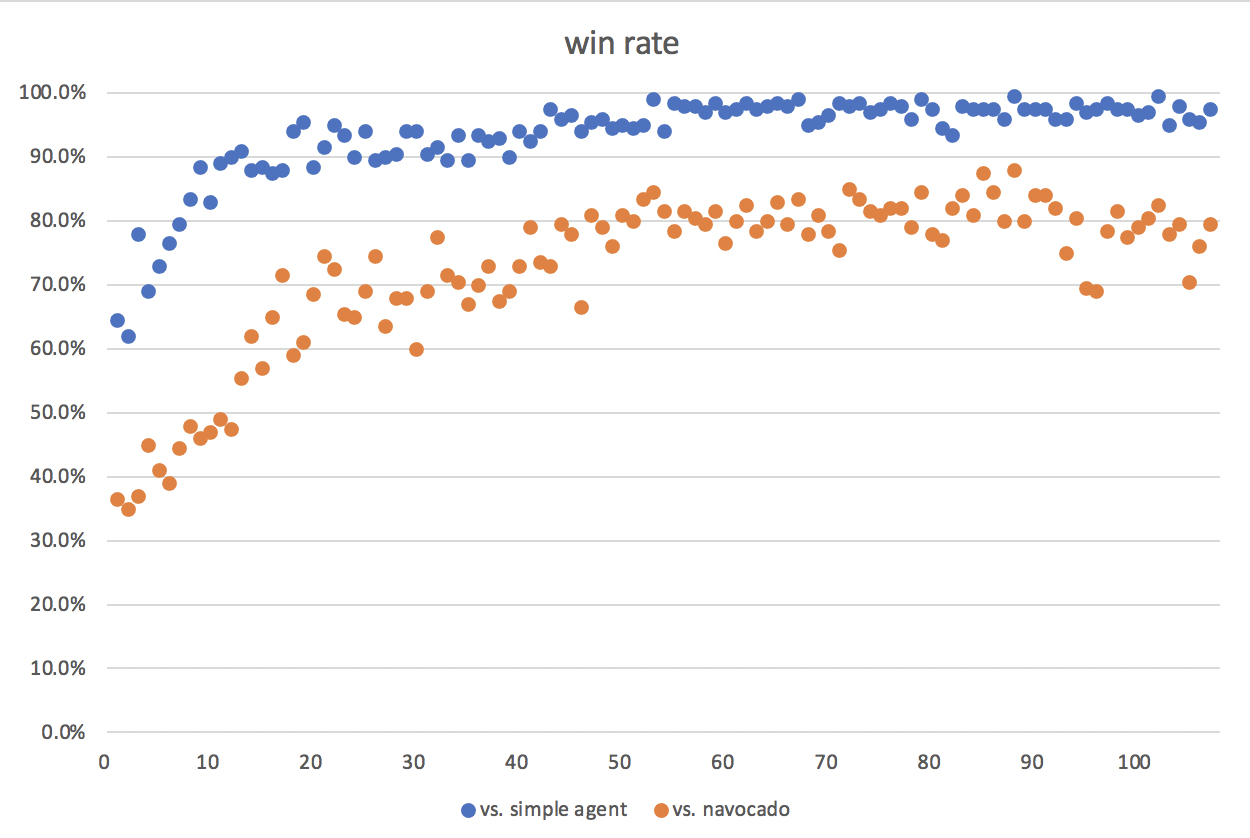}
    \hspace{1em}
    \includegraphics[width=0.45\textwidth]{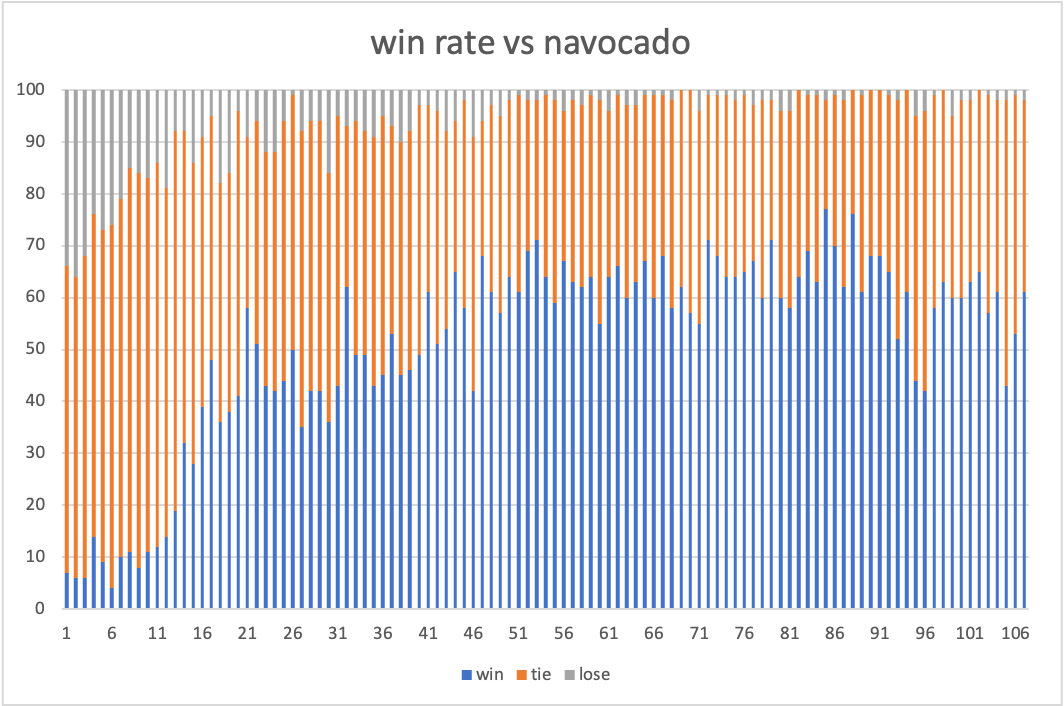}
    \caption{Win-rates curve for the training iteration. Each number is reported by taking 100 games. 
    \textbf{Left}: Win-rates of our agent against Simple Agent, where a tie is counted as 0.5 win and 0.5 lose.
    \textbf{Right}: Number of "wins/losses/ties" for our agents against Navocado. 
    }
    \label{fig:pommerman}
\end{figure}

We test our agents by playing against Simple Agent (a rule-based builtin AI of Pommerman) and Navocado (announced as the best learning-based agent in NeurIPS 2018 Competition). 
The win-rates curves for the training iteration is shown in Fig.~\ref{fig:pommerman}.
As can be seen, 
the trained agent outperforms both Simple Agent and Navocado.

\subsection{Throughput}
\label{sec:exp-throughput}
To help readers evaluate how fast and how large-scale the training requires for a non-trivial environment,
we provide some information in Table~\ref{tab:throughput},
where $M_G$ denotes the number of the parallel learning agents (Section~\ref{sec:design}),
``\#CPU cores'' and ``\#GPUs'' denote how many CPU cores and GPUs are required per learning agent, respectively.
The total number should be multiplied by $M_G$.
For example,
in TStarBot-X we use $96 \times 3 = 192$ GPUs in total.
Unless noted otherwise,
the GPU type is V100 where RDMA/RoCE is disabled and only TCP connection is used for Horovod allreduce.
``rfps'' denotes the receiving frames-per-second (number of frames sent from the Actors),
and ``cfps'' denotes the consuming frames-per-second (number of frames learned on the Learners).
In Table~\ref{tab:throughput}, 
rfps and cfps are also reported as per learning agent.
The ``in-game fps'' means how many frames are rendered by the game core for an in-game second,
for which we've accounted for the frame-skip.
Take ViZDoom for example, 
there are 35 raw frames for one in-game second~\cite{vizdoom-fps}.
In our experiments we use a frame-skip $=2$, 
henceforth the number 17.5 ($=35/2$) in Table~\ref{tab:throughput}.
One can infer the training speed-up or how long the in-game time has been spent by reading the rfps and in-game fps.
When rfps and cfps are almost equal (e.g., implementing a blocking queue for receiving data on the learner),
the on-policyness for an RL algorithm will appear to be good.
When cfps $>$ rfps, the ratio cfps/rfps indicates on average how many times a frame is learned repeatedly.

\begin{threeparttable}[htbp]
  \centering
  \caption{Throughput and other related information for several environments. See the texts for explanations.}
  \label{tab:throughput}
  \small
  \begin{tabular}{l|l|l|l|l|l|l}
    \hline
    Env & $M_G$ & \#CPU cores & \#GPUs & rfps & cfps & in-game fps \\
    \hline\hline
    Dota 2 1v1~\cite{openai-five-blog} & 1 & 60,000 & 256\tnote{a} & 1.1M & 2.8M & 10 \\
    \hline
    Dota 2 5v5~\cite{openai-five-blog} & 1 & 128,000 & 256\tnote{b} & 493K & 1.0M & 7.5 \\
    \hline
    StarCraft II (AlphaStar~\cite{vinyals2019grandmaster})\tnote{*} & 12 & 4,200 & 256\tnote{c} & 25K & 50K & <4.4 \\
    \hline
    Quake III~\cite{jaderberg2019human} & 30  & $64 \times c$\tnote{$\ddagger$} & N/A & N/A & N/A & N/A \\
    \hline\hline
    StarCraft II (TStarBot-X~\cite{han2020tstarbotx})\tnote{$\dagger$} & 3 & 4,200 & 96 & 1.7K & 4.2K & 1.7 \\
    \hline
    ViZDoom & 1 & 1,152 & 32\tnote{d}  & 6.0K & 8.2K & 17.5 \\
    \hline
    Pommerman & 1 & 100 & 2  & 2.9K & 20.0K & N/A \\
    \hline
  \end{tabular}
  \begin{tablenotes}
     \small
     \item[*] All the three races: Terran, Protoss, Zerg
     \item[$\dagger$] Only one race: Zerg
     \item[$\ddagger$] The constant $c$ denotes number of CPU cores assigned to a game core process, which is not reported in~\cite{jaderberg2019human}
     \item[a] K80
     \item[b] P100
     \item[c] The number 256 means 256 TPU-v3 cores
     \item[d] M40
   \end{tablenotes}
\end{threeparttable}

\section{Conclusion and Future Work}
\label{sec:conclusion}
We introduce an open source framework for competitive self-play based Multi-Agent Reinforcement Learning,
referred to as TLeague.
It is able to perform distributed training over a heterogeneous cluster (hybrid of GPU and CPU machines),
achieving a high throughput and a reasonable scale-up.
It can be run over any standard k8s cluster and be deployed on most Cloud Computing vendors (e.g., the Tencent Cloud).
The code is well structured and is easy to extend for your own interested applications. 
Using common policy gradient and opponent sampling algorithms,
we've shown that the agent trained with TLeague yields satisfactory results for several popular benchmark environments including StarCraft II (zvz full game), ViZDoom (CIG 2016 track 1) and Pommerman (NeurIPS 2018 2vs2 competition).
It will be interesting to consider whether the TLeague large-scale training be applicable to even more complicated problems,
e.g., 
the strategic video game CMANO~\cite{cmano},
the trading system~\cite{bao2019multi},
the real-world robotics such as the FPV drone racing~\cite{airsim,drone-racing} and the robot swarm~\cite{Werfel2014Designing,2014Programmable}.

\subsubsection*{Acknowledgement}
Thanks Qing Wang (drwang) for developing an early version of the framework.
Thanks Zhuobin Zheng (jackzbzheng) and Jiaming Lu (loyavejmlu) for initiating the ViZDoom experiments during the internship with Tencent AI Lab.
Thanks 
Yinyuting Yin (mailyyin),
Tengfei Shi (francisshi),
Bei Shi (beishi),
Haobo Fu (haobofu)
and Xipeng Wu (haroldwu) for helpful discussions on distributed training.
Thanks
Qingwei Guo (leoqwguo),
Xinan Jiang (xinanjiang),
Feihu Zhou (hopezhou)
for showing us the advanced usage of Tensorflow.
Thanks 
Huayuan Xiao (howardxiao), 
Ling Cui (gracecui),
Ang Li (marcriverli),
Zhiguo Hong (zhiguohong),
Jike Song (jikesong),
Dekai Li (asinli),
Guangyou Yu (garyyu),
Hui Zou (joezou),
Dandan Song (dandansong),
Keyang Xie (keyangxie),
Hongyu Zou (leonardzou),
Shiyong Wang (warriorwang),
Chong Zha (tillerzha),
Xiaofei Wang (kendywang),
Rennan Yao (liamyao)
and many other colleagues from Tecent CSIG and TEG for preparing and maintaining the compute resources.
Shuxing Li (meowli) and Jiawei Xu (ztjiaweixu) are funded by Tencent AI Lab RhinoBird Focused Research Program JR201986.

\small

\bibliography{cite}
\bibliographystyle{unsrt}

\end{document}